\documentclass[10pt,twocolumn,letterpaper]{article}

\usepackage{iccv}
\usepackage{times}
\usepackage{epsfig}
\usepackage{graphicx}
\usepackage{amsmath}
\usepackage{amssymb}

\usepackage{multirow}
\usepackage{enumitem}
\usepackage{caption}
\usepackage{subcaption}
\usepackage[accsupp]{axessibility}

\usepackage[breaklinks=true,bookmarks=false]{hyperref}

\iccvfinalcopy 

\def\iccvPaperID{8643} 

\ificcvfinal\fi

\graphicspath{{figures/}}

\begin{document}

\title{CanvasVAE: Learning to Generate Vector Graphic Documents}

\author{Kota Yamaguchi\\
CyberAgent\\
{\tt\small yamaguchi\_kota@cyberagent.co.jp}
}

\maketitle
\ificcvfinal\thispagestyle{empty}\fi

\begin{abstract}
Vector graphic documents present visual elements in a resolution free, compact format and are often seen in creative applications.
In this work, we attempt to learn a generative model of vector graphic documents.
We define vector graphic documents by a multi-modal set of attributes associated to a canvas and a sequence of visual elements such as shapes, images, or texts, and train variational auto-encoders to learn the representation of the documents.
We collect a new dataset of design templates from an online service that features complete document structure including occluded elements.
In experiments, we show that our model, named CanvasVAE, constitutes a strong baseline for generative modeling of vector graphic documents.
\end{abstract}

\section{Introduction}
In creative workflows, designers work on visual presentation via vector graphic formats.
2D vector graphics represent images in a compact descriptive structure; instead of spatial array of pixels, graphic documents describe a canvas and arrangement of visual elements such as shapes or texts in a specific format like SVG or PDF.
Vector graphics are crucial in creative production for its resolution-free representation, human interpretability, and editability.
Because of its importance in creative applications, there has been a long but active history of research on tracing vector graphic representation from a raster image~\cite{selinger2003potrace,kopf2011depixelizing,bessmeltsev2019vectorization,liu2019learning,reddy2021im2vec}.

In this work, we study a generative model of vector graphic documents.
While raster-based generative models show tremendous progress in synthesizing high-quality images~\cite{johnson2018image,park2019SPADE,karras2019stylegan2}, there has been relatively scarce studies on vector graphic documents~\cite{zheng2019content,carlier2020deepsvg,li2020attribute}.
Although both raster and vector graphics deal with images, vector graphics do not have canvas pixels and cannot take advantage of the current mainstream approach of convolutional neural networks without rasterization, which is typically not differentiable~\cite{li2020differentiable}.
Learning a generative model of vector graphics therefore imposes us unique challenges in 1) how to represent complex data structure of vector graphic formats in a unified manner, 2) how to formulate the learning problem, and 3) how to evaluate the quality of documents.

We address the task of generative learning of vector graphics using a variational auto-encoder (VAE)~\cite{kingma2013auto}, where we define documents by a multi-modal combination of canvas attributes and a sequence of element attributes.
Unlike conditional layout inference~\cite{zheng2019content,lee2020neural,li2020attribute}, we consider unconditional document generation including both a canvas and variable number of elements.
Our architecture, named CanvasVAE, learns an encoder that projects a given graphic document into a latent code, and a decoder that reconstructs the given document from the latent code.
We adopt Transformer-based network~\cite{devlin2018bert} in both the encoder and the decoder to process variable-length sequence of elements in a document.
The learned decoder can take randomly sampled latent code to generate a new vector graphic document.
For our study, we collect a large-scale dataset of design templates for our study that offers complete document structure and content information.
In evaluation, we propose to combine normalized metrics for all attributes to measure the overall quality of reconstruction and generation.
We compare several variants of CanvasVAE architecture and show that a Transformer-based model constitutes a strong baseline for the vector graphic generation task.

We summarize our contributions in the following.
\begin{enumerate}[itemsep=-.7ex]
    \item We propose the CanvasVAE architecture for the task of unconditional generative learning of vector graphic documents, where we model documents by a structured, multi-modal set of canvas and element attributes.
    \item We build Crello dataset, which is a dataset consisting of large number of design templates and features complete vector information including occluded elements.
    \item We empirically show that our Transformer-based variant of CanvasVAE achieves a strong performance in both document reconstruction and generation.
\end{enumerate}

\section{Related work}

\paragraph{Generative layout modeling}
There has been several attempts at conditional layout modeling where the goal is to generate bounding box arrangements given certain inputs.
LayoutVAE~\cite{jyothi2019layoutvae} learns a two-stage autoregressive VAE that
takes a label set and generates bounding boxes for each label, for scene image representation.
For design applications, Zheng \etal~\cite{zheng2019content} report a generative model for magazine layout conditioned on a set of elements and meta-data, where raster adversarial networks generate layout maps.
Lee \etal~\cite{lee2020neural} propose a three-step approach to predict a layout given an initial set of elements that accepts partial relation annotation.
Li \etal~\cite{li2019layoutgan,li2020attribute} learn a model that refines the geometry of the given elements, such that the refined layout looks realistic to a discriminator built on a differentiable wire-frame rasterizer.
Tan \etal~\cite{tan2019text2scene} propose text-to-scene generation that explicitly considers a layout and attributes.
Wang \etal~\cite{wang2020learning} consider a reinforcement learning approach to select appropriate elements for the given document.
For UI layout domain, Manandhar \etal~\cite{manandhar2020learning} propose to learn UI layout representation by metric learning and raster decoder.
Li \etal~\cite{li2021screen2vec} recently report an attempt in multi-modal representation learning of UI layout.

In contrast to conditional layout generation, we tackle on the task of \emph{unconditional} document generation including layout and other attributes.
Gupta \etal recently report autoregressive model for generating layout~\cite{gupta2020layout} but without learning a latent representation and instead relies on beam search.
READ~\cite{patil2020read} is the only pilot study similar to our unconditional scenario, although their recursive model only considers labeled bounding box without content attributes.
Arroyo \etal~\cite{arroyo2021variational} very recently report a layout generation model.
Our model fully works in symbolic vector data without explicit rasterization~\cite{zheng2019content,li2019layoutgan}, which allows us to easily process data in a resolution free manner.

\paragraph{Vector graphic generation}
Although our main focus is document-level generation, there has been several important work in stroke or path level vector graphic modeling that aims at learning to generate resolution-free shapes.
Sketch RNN is a pioneering work on learning drawing strokes using recurrent networks~\cite{ha2017neural}.
SPIRAL is a reinforcment adversarial learning approach to vectorize a given raster image~\cite{ganin2018synthesizing}.
Lopes \etal learn an autoregressive VAE to generate vector font strokes~\cite{lopes2019learned}.
Song \etal report a generative model of B{\'e}zier curves for sketch strokes~\cite{song2020beziersketch}.
Carlier \etal propose DeepSVG architecture that consists of a hierarchical auto-encoder that learns a representation for a set of paths~\cite{carlier2020deepsvg}.
We get many inspirations from DeepSVG especially in our design of oneshot decoding architecture.


\begin{table*}[t]
\centering
\footnotesize
\caption{Attribute descriptions for vector graphic data}\label{tab:columns}
\begin{tabular}{|l|l|lllll|}
\hline
Dataset & Attribute of & Name          & Type        & Size  & Dim & Description                                                           \\
\hline
\multirow{12}{*}{Crello}  & \multirow{6}{*}{Canvas} & Length & Categorical & 50 & 1 & Length of elements up to 50 \\
  && Group         & Categorical &  7 & 1         & Broad design group, such as social media posts or blog headers \\
  && Format        & Categorical &  68 & 1        & Detailed design format, such as Instagram post or postcard \\
&& Width         & Categorical &  42 & 1        & Canvas pixel width available in crello.com                            \\
&& Height        & Categorical &  47 & 1        & Canvas pixel height available in crello.com                           \\
&& Category      & Categorical &  24 & 1        & Topic category of the design, such as holiday celebration             \\
\cline{2-7}
& \multirow{6}{*}{Element} & Type          & Categorical & 6 & 1 & Element type, such as vector shape, image, or text placeholder     \\
&& Position      & Categorical &  64 & 2        & Left and top position each quantized to 64 bins \\
&& Size          & Categorical &  64 & 2        & Width and height each quantized to 64 bins \\
&& Opacity       & Categorical &  8 & 1         & Opacity quantized to 8 bins \\
&& Color         & Categorical &  16 & 3        & RGB color each quantized to 16 bins, only relevant for solid fill and texts \\
&& Image & Numerical   & 1 & 256        & Pre-trained image feature, only relevant for shapes and images \\
\hline
\multirow{7}{*}{RICO} & Canvas       & Length        & Categorical &  50 & 1        & Length of elements up to 50                                           \\
\cline{2-7}
& \multirow{6}{*}{Element} & Component     & Categorical &  27 & 1       & Element type, such as text, image, icon, etc. \\
&& Position      & Categorical &  64 & 2        & Left and top position each quantized to 64 bins \\
&& Size          & Categorical &  64 & 2        & Width and height each quantized to 64 bins \\
&& Icon          & Categorical &  59 & 1        & Icon type, such as \emph{arrow}, \emph{close}, \emph{home} \\
&& Button   & Categorical &  25 & 1             & Text on button, such as \emph{login} or \emph{back} \\
&& Clickable     & Categorical &  2 & 1        & Binary flag indicating if the element is clickable \\
\hline
\end{tabular}
\end{table*}
\section{Vector graphic representation} \label{sec:vector-graphic-representation}

\subsection{Document structure}
In this work, we define vector graphic documents to be a single-page canvas and associated sequence of visual elements such as texts, shapes, or raster images.
We represent a document $X = (X_c, X_E)$ by a set of canvas attributes $X_c  = \{ \mathbf{x}_k | k \in \mathcal{C}\}$ and a sequence of elements $X_E = \{ X_e^1, X_e^2, \cdots, X_e^T \}$, where $X_e^t = \{ \mathbf{x}_k^t | k \in \mathcal{E}\}$ is a set of element attributes.
We denote a set of canvas and element attribute indices by $\mathcal{C}$ and $\mathcal{E}$, respectively.
Canvas attributes represent global document properties, such as canvas size or document category.
Element attributes indicate element-specific configuration, such as position, size, type of the element, opacity, color, or a texture image if the element represents a raster image.
In addition, we explicitly include the element length in the canvas attributes $X_c$.
We represent elements by a sequence, where the order reflects the depth of which elements appear on top.
The actual attribute definition depends on datasets we describe in the next section.

\subsection{Datasets}

\paragraph{Crello dataset}\label{sec:crello}
Crello dataset consists of design templates we obtained from online design service, \emph{crello.com}.
The dataset contains designs for various display formats, such as social media posts, banner ads, blog headers, or printed posters, all in a vector format.
In dataset construction, we first downloaded design templates and associated resources (e.g., linked images) from \emph{crello.com}. 
After the initial data acquisition, we inspected the data structure and identified useful vector graphic information in each template.
Next, we eliminated mal-formed templates or those having more than 50 elements, and finally obtained 23,182 templates.
We randomly partition the dataset to 18,714 / 2,316 / 2,331 examples for train, validation, and test splits.

In Crello dataset, each document has canvas attributes $X_c$ = \{$\mathbf{x}_\mathrm{length}$, $\mathbf{x}_\mathrm{width}$, $\mathbf{x}_\mathrm{height}$, $\mathbf{x}_\mathrm{group}$, $\mathbf{x}_\mathrm{category}$, $\mathbf{x}_\mathrm{format}$\} and element attributes $X_e^t$ = \{$\mathbf{x}_\mathrm{type}^t$, $\mathbf{x}_\mathrm{position}^t$, $\mathbf{x}_\mathrm{size}^t$, $\mathbf{x}_\mathrm{color}^t$, $\mathbf{x}_\mathrm{opacity}^t$, $\mathbf{x}_\mathrm{image}^t$\}.
Table \ref{tab:columns} summarizes the detail of each attribute.
Image and color attributes are exclusive; we extract color for text placeholders and solid backgrounds, and we extract image features for shapes and image elements in the document.
Except for image features, we quantize numeric attributes to one-hot representations, such as element position, size, or colors, because 1) discretization implicitly enforces element alignment, and 2) attributes often do not follow normal distribution suitable for regression.
The image feature allows us content-aware document modeling, and also is useful for visualization purpose.

We obtain image features using a raster-based convolutional VAE that we pre-train from all the image and shape elements in the Crello dataset.
We do not use ImageNet pre-trained model here, because ImageNet does not contain alpha channels nor vector shapes.
For pre-training of the VAE, we rasterize all the image and shape elements in $256 \times 256$ pixel canvas with resizing, and saves in RGBA raster format.
From the rasterized images, we learn a VAE consisting of MobileNetV2-based encoder~\cite{sandler2018mobilenetv2} and a 6-layer convolutional decoder.
After pre-training the convolutional VAE, we obtain a 256-dimensional latent representation using the learned image encoder for all the image and shape elements in Crello dataset.

In contrast to existing layout datasets that mainly consider a set of labeled bounding boxes~\cite{deka2017rico,zheng2019content,zhong2019publaynet}, our Crello dataset offers complete vector graphic structure including appearance for occluded elements.
This enables us to learn a generative model that considers the appearance and attributes of graphic elements in addition to the layout structure. Also, Crello dataset contains canvas in various aspect ratio.
This imposes us a unique challenge, because we have to handle variable-sized documents that raster-based models do not work well with.

\paragraph{RICO dataset}\label{sec:rico}
RICO dataset offers a large number of user interface designs for mobile applications with manually annotated elements~\cite{deka2017rico}.
We use RICO dataset to evaluate the generalization ability of our CanvasVAE.
All the UI screenshots from RICO have a fixed resolution of $2560 \times 1440$ pixels, and there is no document-wise label.
We set canvas attributes to only have element length: $X_c$ = \{$\mathbf{x}_\mathrm{length}$\}, and for each element, we model $X_e^t$ = \{$\mathbf{x}_\mathrm{component}^t$, $\mathbf{x}_\mathrm{position}^t$, $\mathbf{x}_\mathrm{size}^t$, $\mathbf{x}_\mathrm{icon}^t$, $\mathbf{x}_\mathrm{button}^t$, $\mathbf{x}_\mathrm{clickable}^t$\}.
Most of the pre-processing follows Crello dataset; we quantize numeric attributes to one-hot representations.
Table \ref{tab:columns} summarizes the attributes we consider in this work.
\begin{figure*}[t]
    \centering
    \includegraphics[width=.9\textwidth]{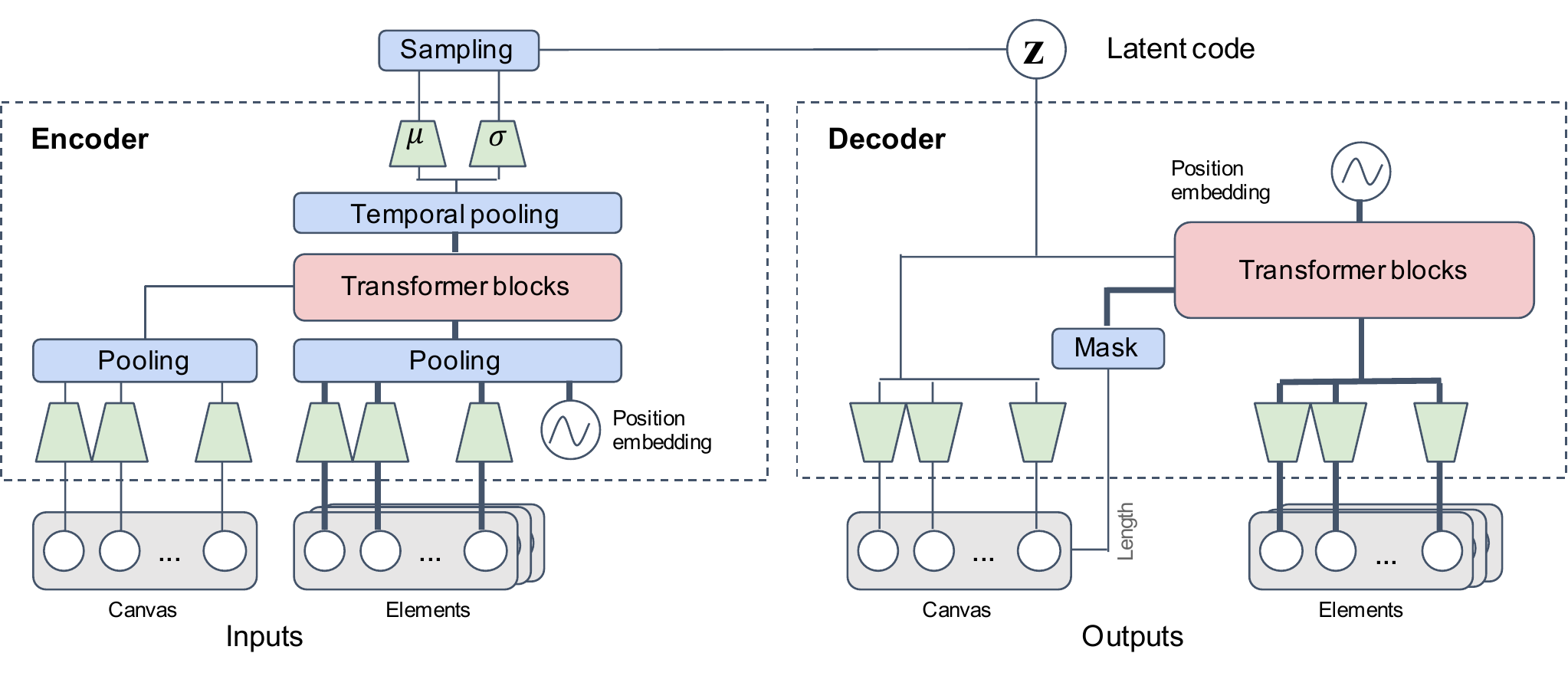}
    \caption{CanvasVAE architecture.}
    \label{fig:architecture}
\end{figure*}

\section{CanvasVAE}\label{sec:canvasvae}
Our goal is to learn a generative model of vector graphic documents.
We aim at learning a VAE that consists of a probabilistic encoder and a decoder using neural networks.

\paragraph{VAE basics}
Let us denote a vector graphic instance by $X$ and a latent code by $\mathbf{z}$.
A VAE learns a generative model $p_\theta(X, \mathbf{z}) = p_\theta(X | \mathbf{z}) p_\theta(\mathbf{z})$ and an approximate posterior $q_\phi(\mathbf{z} | X)$, using variational lower bounds~\cite{kingma2013auto}:
\begin{align}
    \mathcal{L}(X; \theta, \phi) =
    & {\mathbb{E}}_{q_\phi (\mathbf{z} | X)} \left[ \log p_\theta ( X | \mathbf{z} ) \right] \nonumber \\
    & - \mathrm{KL}(q_\phi (\mathbf{z} | X) || p_\theta (\mathbf{z})),  \label{eq:elbo}
\end{align}
where $\phi$ is the parameters of the inference model and $\theta$ is the parameters of the generative model.

We model the approximate variational posterior $q_\phi(\mathbf{z} | X)$ by a Gaussian distribution with a diagonal covariance:
\begin{align}
    q_\phi(\mathbf{z}|X)
    &\equiv \mathcal{N}(\mathbf{z}; \mathbf{\mu}_{\phi}(X), \mathbf{\sigma}_{\phi}(X)^2 \mathbf{I}), \label{eq:variational_posterior}
\end{align}
where $\mathbf{\mu}_{\phi}(X)$ and $\mathbf{\sigma}_{\phi}(X)$ are the encoder outputs that we model by a neural network.
We set the prior over the latent code $\mathbf{z}$ to be a unit multivariate Gaussian $p_\theta(\mathbf{z}) \equiv \mathcal{N}(\mathbf{z}; \mathbf{0}, \mathbf{I})$.
We also model the data likelihood $p_\theta(X | \mathbf{z})$ using a neural network.
Fig~\ref{fig:architecture} illustrates our CanvasVAE encoder and decoder architecture.

\paragraph{Encoder}\label{sec:encoder}
Our encoder takes a vector graphic input $X$ and predicts the parameters of approximate prior $\mu_\phi(X)$, $\sigma_\phi(X)$.
We describe the encoder in the following:
\begin{align}
    \mathbf{h}_{c} &= \textstyle{\sum}_{k \in \mathcal{C}} f_{k}( \mathbf{x}_k; \phi), \\
    \mathbf{h}_{e}^t &= \textstyle{\sum}_{k \in \mathcal{E}} f_k(\mathbf{x}_k^t; \phi) + \mathbf{x}_\mathrm{position, \phi}^t, \\
    \mathbf{h}_\mathrm{enc} &= \frac{1}{T} \textstyle{\sum}_t^T B(\{ \mathbf{h}_{e}^t \}, \mathbf{h}_{c}; \phi), \label{eq:encoder-seq-pooling} \\
    \mu_\phi(X) &= f_\mu (\mathbf{h}_\mathrm{enc}; \phi), \\
    \sigma_\phi(X) &= f_\sigma (\mathbf{h}_\mathrm{enc}; \phi).
\end{align}
The encoder first projects each canvas attribute $\mathbf{x}_k$ using a feed-forward layer $f_k$ to the dimensionality, and adds up to make the hidden side input $\mathbf{h}_c$ to the Transformer block.
Similarly, the encoder projects each element attribute $\mathbf{x}_k^t$ using a feed-forward layer to the same dimensionality, and adds up together with the position embedding $\mathbf{x}_\mathrm{position, \phi}^t$ to make the hidden input $\mathbf{h}_e^t$ to the Transformer block for each step $t$.
The positional embedding $\mathbf{x}_\mathrm{position, \phi}^t$ provides information on absolute position within the element sequence~\cite{devlin2018bert}.
We learn the positional embedding during training.
$B$ is a variant of Transformer model that adds a side input between the self attention and the feed-forward layer, which is similar to the decoder block of DeepSVG~\cite{carlier2020deepsvg}.
We stack up multiple Transformer blocks to transform input embedding $\mathbf{h}_{c}$ and $\mathbf{h}_{e}^t$ to produce temporally pooled internal representation $\mathbf{h}_\mathrm{enc}$.
$f_\mu$ and $f_\sigma$ are the last feed-forward layer of the encoder to produce $\mu_\phi(X)$ and $\sigma_\phi(X)$.

\paragraph{Decoder}\label{sec:decoder}
Our decoder takes a sampled latent code $\mathbf{z}$ and produces reconstruction $\hat{X}$ from $p_\theta(X | \mathbf{z}) = \prod_k p_\theta (\mathbf{x}_k| \mathbf{z}) \prod_t p_\theta (\mathbf{x}_k^t| \mathbf{z})$.
We describe our decoder by:
\begin{align}
    \mathbf{h}_\mathrm{dec}^t & = B(\{\mathbf{x}_\mathrm{position,\theta}^t\}, \mathbf{z}; \theta), \label{eq:decoder-seq} \\
    p_\theta(\mathbf{x}_k | \mathbf{z}) &= f_{k}(\mathbf{x}_k, \mathbf{z}; \theta), \\
    p_\theta(\mathbf{x}_k^t | \mathbf{z}) &= f_{k}(\mathbf{x}_k^t, \mathbf{h}_\mathrm{dec}^t; \theta),
\end{align}
where $f_k$ is the last feed-forward network for the attribute $k$.
Our decoder uses the same Transformer block $B$ with the encoder.
The decoder has the positional embedding $\mathbf{x}_\mathrm{position, \theta}^t$ to feed the absolute position information for sequence reconstruction.

At generation, we apply stochastic sampling to $\mathbf{z}$ and obtain a maximum likelihood estimate from our decoder head $p_\theta(\mathbf{x}_k|\mathbf{z})$ for categorical attributes, or regression output for numerical attributes.
To generate a sequence from the latent code $\mathbf{z}$, we have to first decide the number of elements in the document.
We predict the length $T$ from $p_\theta(\mathbf{x}_\mathrm{length}|\mathbf{z})$, and feed the masking information to the Transformer block to exclude out-of-sequence elements in self-attention, and drop extra elements at the final reconstruction.

\paragraph{Loss function}\label{sec:loss}
We derive the loss function for our CanvasVAE from the variational lower bounds (Eq~\ref{eq:elbo}).
For a sample $X$ in our dataset, the loss for each document is given by:
\begin{align}
    \mathcal{L}(X, \hat{X}; \theta, \phi)
    &= \sum_{k \in \mathcal{C}} \mathcal{L}_{k}(\mathbf{x}_k, \hat{\mathbf{x}}_k)
    + \sum_{k \in \mathcal{E}} \sum_{t}^T \mathcal{L}_{k}(\mathbf{x}_k^t, \hat{\mathbf{x}}_k^t) \nonumber \\
    & + \lambda_\mathrm{KL} \mathrm{KL}(\mathcal{N}(\mathbf{z}; \mu_\phi, \sigma_\phi) || \mathcal{N}(\mathbf{z}; \mathbf{0}, \mathbf{I})) \nonumber \\
    & + \lambda_\mathrm{L2} (| \phi |^2 + | \theta |^2),
    \label{eq:loss}
\end{align}
where $\lambda_\mathrm{KL}$ and $\lambda_\mathrm{L2}$ are hyper-parameters to weight the regularization terms.
$\mathcal{L}_{k}$ is a loss term for attribute $k$.
We use cross entropy for categorical attributes and mean squared error for numeric attributes.
At training time, we use teacher-forcing; we discard the predicted length $\hat{T}$ and force the ground truth length $T$ in the decoder.

\section{Experiments} \label{sec:experiments}

We evaluate our CanvasVAE in reconstruction and generation scenarios.
In reconstruction, we evaluate the overall capability of our encoder-decoder model to reproduce the given input.
In generation scenario, we evaluate the decoder capability in terms of the quality of randomly generated documents.

\subsection{Evaluation metrics} \label{sec:evaluation_metrics}
\subsubsection{Reconstruction metrics} \label{sec:reconstruction_metrics}
We have to be able to measure the similarity between two documents to evaluate reconstruction quality.
Unlike raster images, there is no standard metric to measure the distance between vector graphic documents.
Our loss function (Eq~\ref{eq:loss}) is also not appropriate due to teacher-forcing of sequence length.
Considering the multi-modal nature of vector graphic formats, we argue that an ideal metric should be able to evaluate the quality of all the modalities in the document at a uniform scale, and that the metric can handle variable length structure.
We choose the following two metrics to evaluate the document similarity.

\paragraph{Structural similarity}
For document $X_1$ and $X_2$, we measure the structural similarity by the mean of normalized scores for each attribute $k$:
\begin{align}
    & S(X_1, X_2) = \nonumber \\
    & \frac{1}{|\mathcal{C}'| + |\mathcal{E}|} \left[
    \sum_{k \in \mathcal{C}'} s_k(\mathbf{x}_{k, 1}, \mathbf{x}_{k, 2})
    + \sum_{k \in \mathcal{E}} s_k(\{\mathbf{x}_{k, 1}^t\}, \{\mathbf{x}_{k, 2}^t\})
    \right] \label{eq:score},
\end{align}
where $s_k \in \left[ 0, 1 \right]$ is a scoring function, and $\mathcal{C}' = \mathcal{C} \backslash \{ \mathrm{length} \}$.
We exclude length from the canvas attributes because element scores take length into account.
For canvas attributes, we adopt \emph{accuracy} as the scoring function since there are only categorical attributes in our datasets.

For categorical element attributes, a scoring function must be able to evaluate variable length elements. We use BLEU score~\cite{papineni2002bleu} that is a precision-based metric often used in machine translation.
BLEU penalizes a shorter prediction by the brevity term: $\exp \left( \min \left(0, 1 - \frac{|T|}{|\hat{T}|}\right) \right)$, where $\hat{T}$ is the predicted element length.
We use unigram BLEU for evaluation, because vector graphic elements do not exhibit strong ordinal constraints and elements can be swapped as long as they do not overlap.
For the image feature that is the only numerical element attribute in Crello, we use the cosine similarity in [0, 1] scale between the average-pooled features over sequence, multiplied by the brevity term of BLEU score.
Note that our structural similarity is not symmetric because BLEU is not symmetric.

In Crello, the presence of \emph{image} and \emph{color} attributes depend on the element \emph{type}, and $\{\mathbf{x}_{k}^t\}$ can become empty.
We exclude empty attributes from $\mathcal{E}$ in the calculation of eq~\ref{eq:score} if either $X_1$ or $X_2$ include empty attributes.

We evaluate reconstruction performance by the average score over the document set $\mathcal{X} = \{ X_i \}$:
\begin{align}
    S_\mathrm{reconst}(\mathcal{X}) = \frac{1}{|\mathcal{R}|} \sum_{i} S(X_i, \hat{X_i}).
    \label{eq:reconst_eval}
\end{align}

\begin{figure*}[t]
    \centering
    \includegraphics[width=\textwidth]{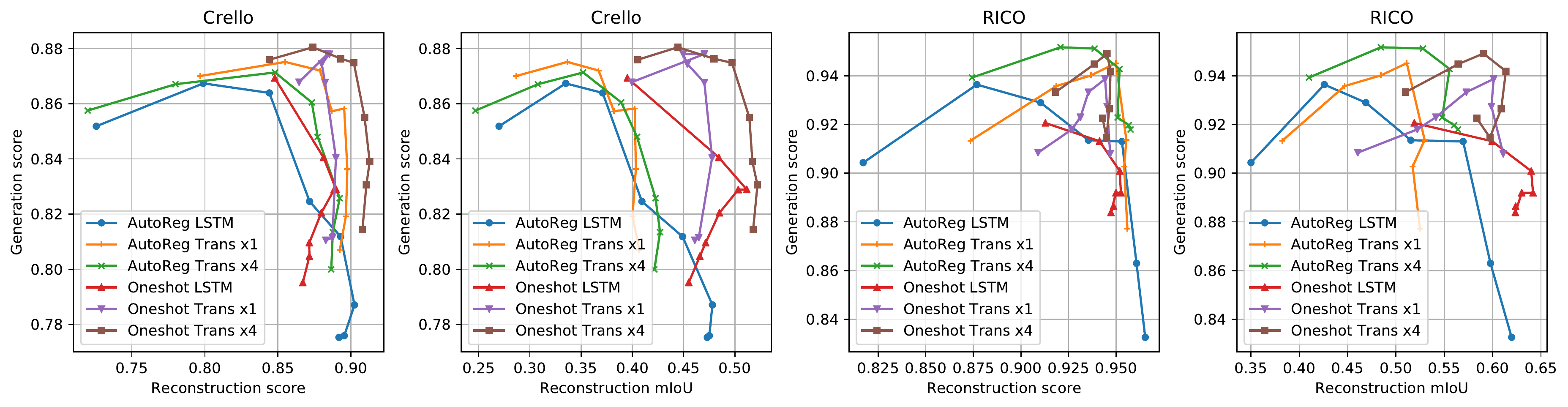}
    \caption{Performance curves in terms of $S_\mathrm{reconst}$ vs. $S_\mathrm{gen}$ and mIoU vs. $S_\mathrm{gen}$ over $\lambda_\mathrm{KL}$ in validation splits. Top-right models show better performance in both reconstruction and generation.}
    \label{fig:performance-curve}
\end{figure*}

\paragraph{Layout mean IoU}
We also include evaluation by mean intersection over union (mIoU) on labeled bounding boxes~\cite{manandhar2020learning} to analyze layout quality.
We use \emph{type} attribute in Crello dataset and \emph{component} attribute in RICO dataset as a primary label for elements.
To compute mIoU, we draw bounding boxes on a canvas in the given element order, compute the IoU for each label, then average over labels.
Since we quantize position and size of each element, we draw bounding boxes on a $64\times64$ grid canvas.
Similar to Eq~\ref{eq:reconst_eval}, we obtain the final score by dataset average.

The mIoU metric ignores attributes other than element position, element size, and element label.
Content attributes such as image or color have no effect on the mIoU metric.
Although Crello dataset has variable-sized canvas, we ignore the aspect ratio of the canvas and only evaluate on the relative position and size of elements.

\subsubsection{Generation metric} \label{sec:generation_metric}
Similar to reconstruction, there is no standard approach to evaluate the similarity between sets of vector graphic documents.
It is also not appropriate to use a raster metric like FID score~\cite{heusel2017gans} because our document can not be rasterized to a fixed resolution nor is a natural image.
We instead define the following metric to evaluate the distributional similarity of vector graphic documents.

For real and generated document sets $\mathcal{X}_1$ and $\mathcal{X}_2$, we first obtain descriptive statistics for each attribute $k$, then compute the similarity between the two sets:
\begin{align}
    S_\mathrm{gen}(\mathcal{X}_1, \mathcal{X}_2) = 
    \frac{1}{|\mathcal{C}| + |\mathcal{E}|} 
    \sum_{k \in \mathcal{C} \cup \mathcal{E}}
    d_k(a_k (\mathcal{X}_{1}), a_k(\mathcal{X}_{2})),
    \label{eq:gen_score}
\end{align}
where $a_k$ is an aggregation function that computes descriptive statistics of attribute $k$, and $d_k$ is a scoring function.
For categorical attributes, we use histogram for $a_k$ and normalized histogram intersection for $d_k$.
For numerical attributes, we use average pooling for $a_k$ and cosine similarity for $d_k$.

\begin{figure*}[t]
\centering
\includegraphics[width=\textwidth]{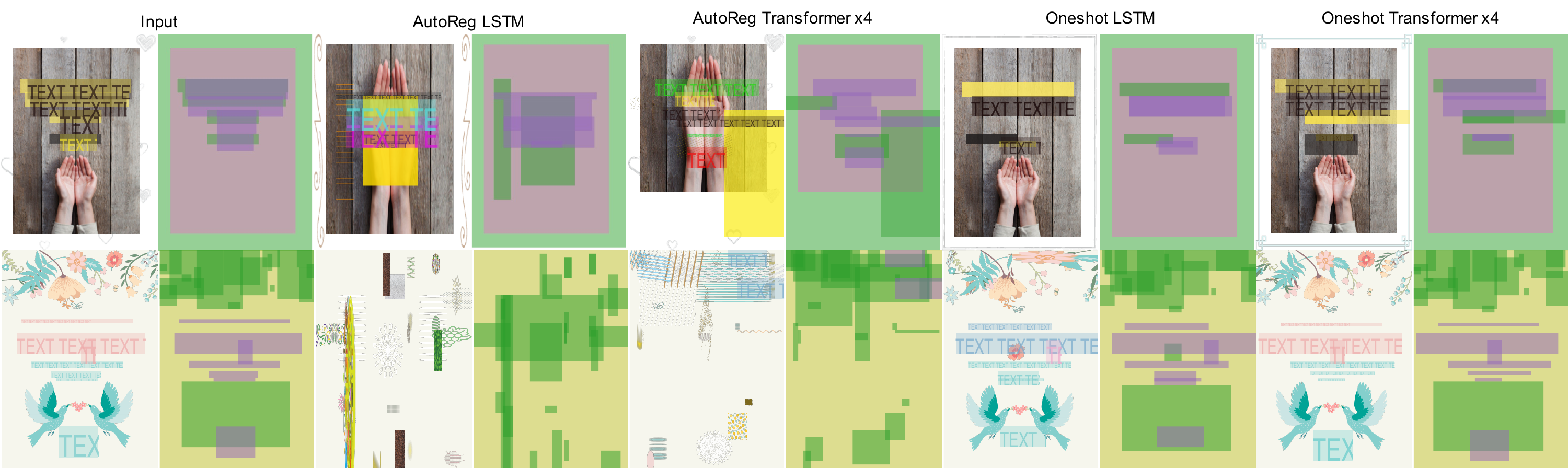}
\caption{Crello reconstruction comparison. For each item, a left picture shows visualization of the design template with colored text placeholders and textured elements, and a right color map illustrates element types. The legend of types are the following: green = \emph{vector shape}, magenta = \emph{image}, purple = \emph{text placeholder}, and yellow = \emph{solid fill}.}
\label{fig:reconstruction-crello}
\end{figure*}
\begin{figure}[t]
\centering
\includegraphics[width=\columnwidth]{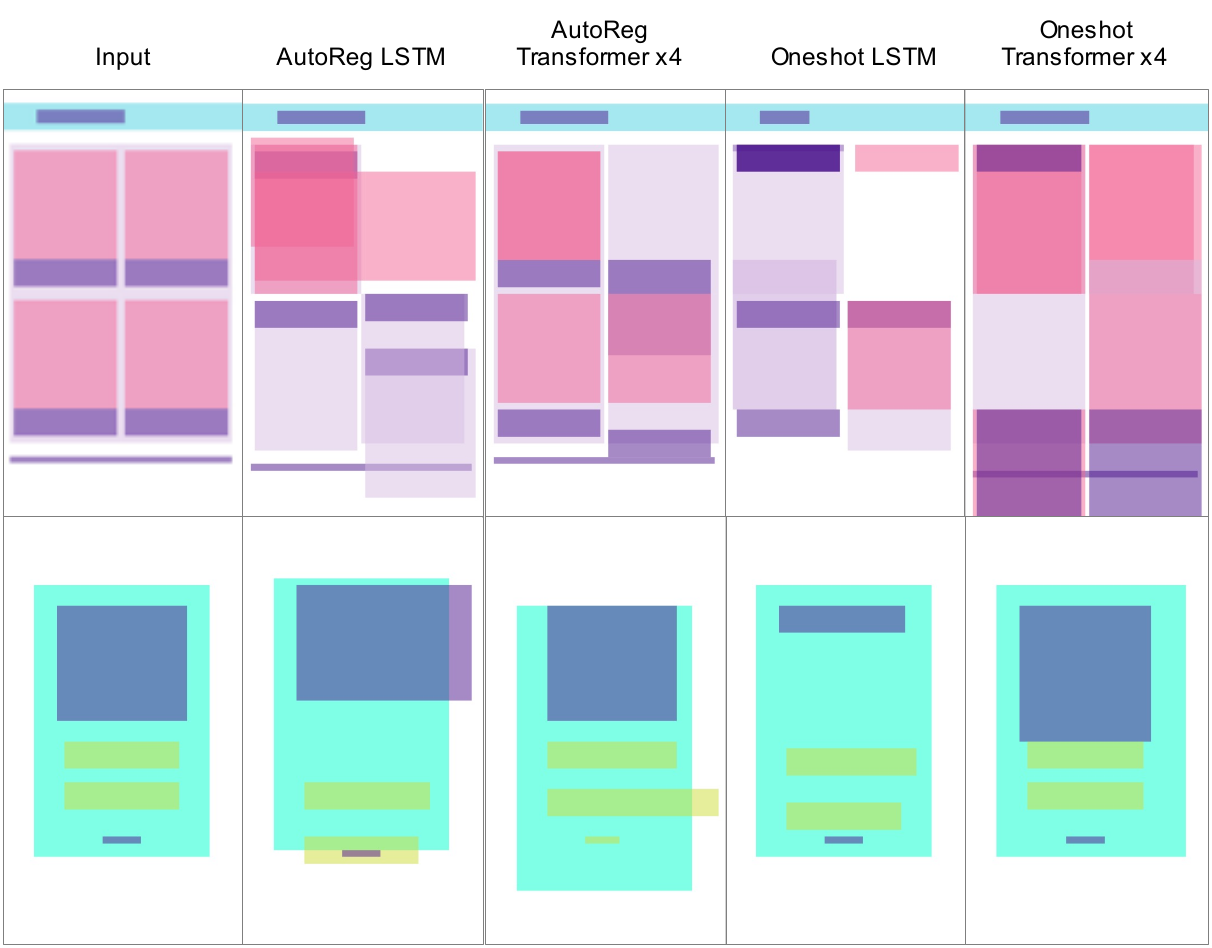}
\caption{RICO reconstruction comparison. Color indicates the component label of each element. Color is from the original RICO schema~\cite{deka2017rico}.}
\label{fig:reconstruction-rico}
\end{figure}

\begin{figure}[t]
\centering
\includegraphics[width=\columnwidth]{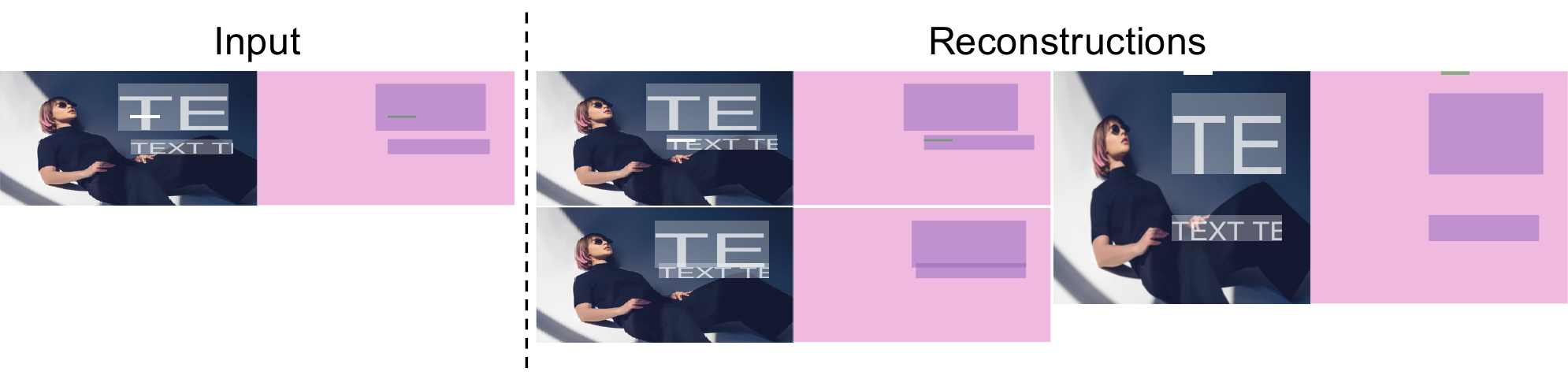}
\caption{Stochastic sampling and reconstruction examples.}
\label{fig:sampling-crello}
\end{figure}
\begin{figure*}[t]
\centering
\includegraphics[width=\textwidth]{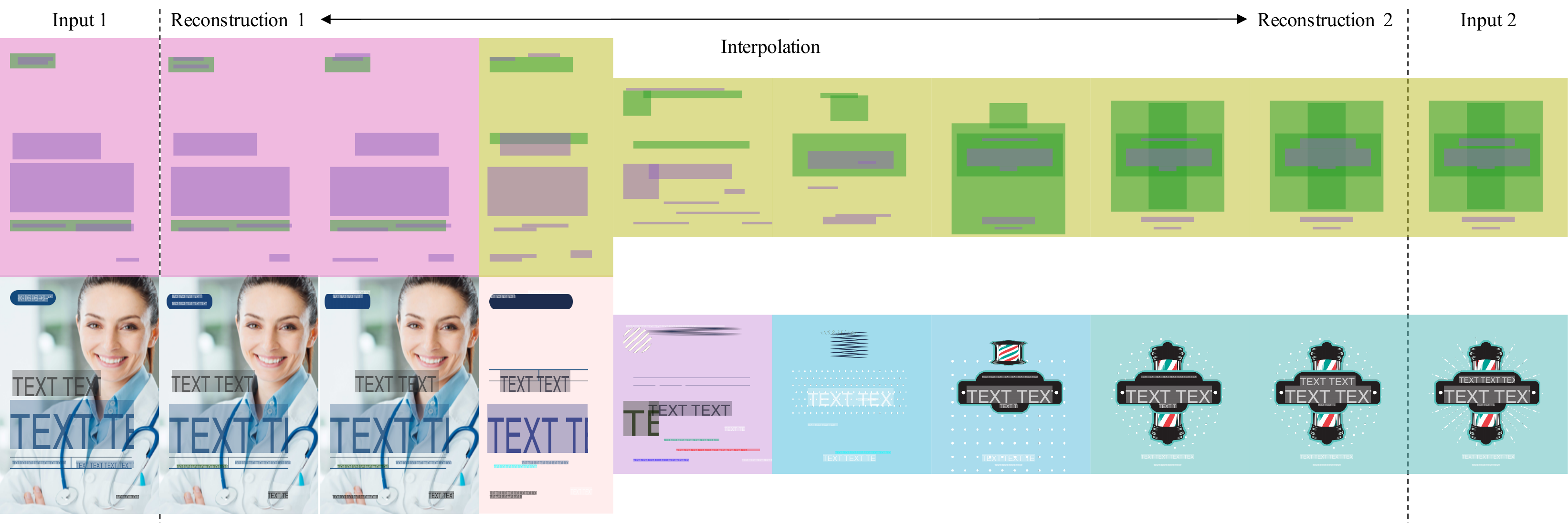}
\caption{Interpolation example.}
\label{fig:interpolation}
\end{figure*}

\begin{figure*}[t]
\centering
\includegraphics[width=\textwidth]{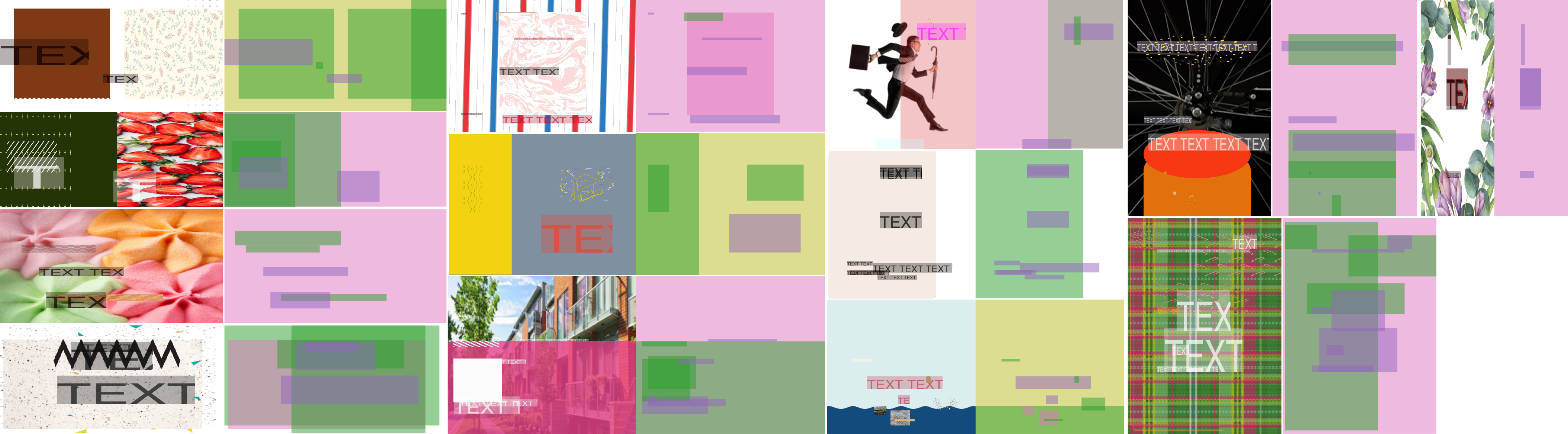}
\caption{Randomly generated Crello documents.}
\label{fig:generation}
\end{figure*}
\begin{figure}[t]
\centering
\includegraphics[width=\columnwidth]{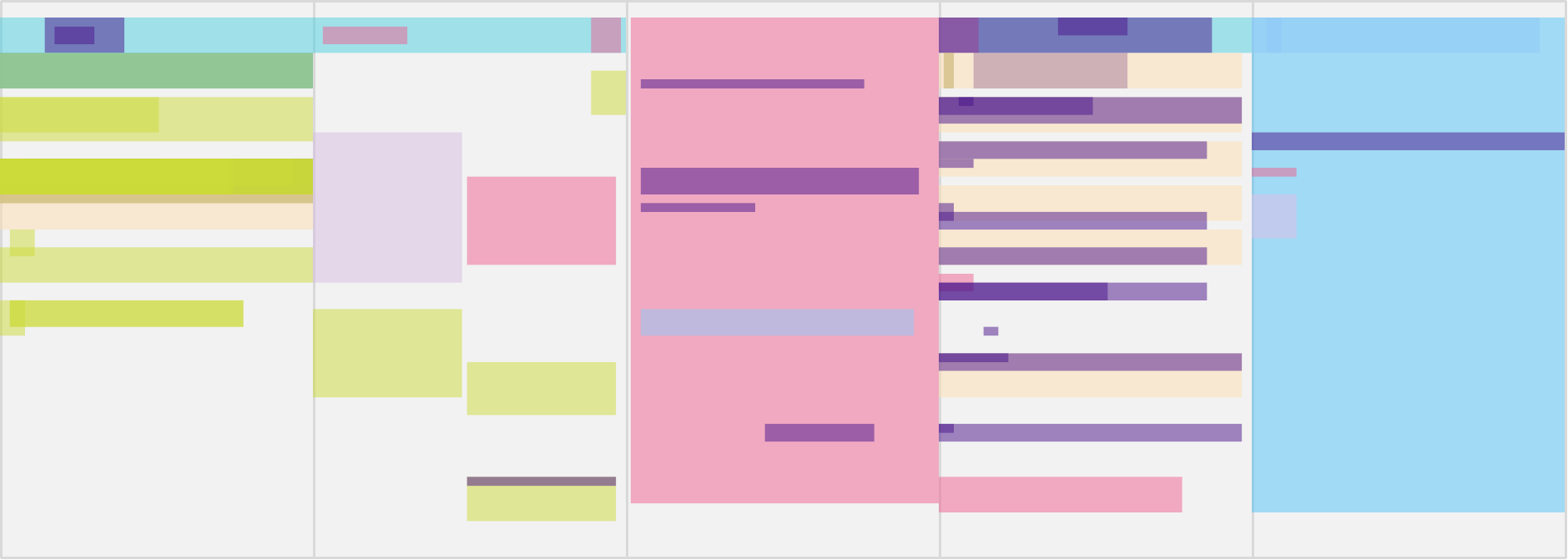}
\caption{Randomly generated RICO documents.}
\label{fig:generation-rico}
\end{figure}

\subsection{Baselines}
We include the following variants of our CanvasVAE as baselines.
Since there is no reported work that is directly comparable to CanvasVAE, we carefully pick comparable building blocks from existing work.

\noindent\textbf{AutoReg LSTM}~ We replace Transformer blocks and temporal pooling in our model (Eq~\ref{eq:encoder-seq-pooling}, \ref{eq:decoder-seq}) with an LSTM.
The side input to the transformer blocks is treated as initial hidden states.
Also, we introduce autoregressive inference procedure instead of our one-shot decoding using positional embedding; we predict an element at $t$ given the elements until $t-1$ in the decoder.
The initial input is a special beginning-of-sequence embedding, which we learn during training.
Our autoregressive LSTM baseline has a decoder similar to LayoutVAE~\cite{jyothi2019layoutvae}, although we do not stochastically sample at each step nor have conditional inputs but have an encoder for the latent code.

\noindent\textbf{AutoReg Transformer}~ Similar to Autoregressive LSTM, we introduce the autoregressive inference but use Transformer blocks.
The decoding process is similar to LayoutTransformer~\cite{gupta2020layout}, but we add an encoder for the latent code.
We also explicitly predict sequence length instead of an end-of-sequence flag to terminate the prediction.

\noindent\textbf{Oneshot LSTM}~ We use positional embedding but replace Transformer blocks with an LSTM.
We use a bidrectional LSTM for this one-shot model because positional embedding allows both past and future information for prediction.

\noindent\textbf{Oneshot Transformer}~ Our CanvasVAE model described in Sec~\ref{sec:canvasvae}.
We also compare the number of Transformer blocks at 1 and 4 for ablation study.

\subsection{Quantitative evaluation}
\begin{table}[t]
    \centering
    \caption{Test performance (\%).}
    \small
    \label{tab:performance}
    \begin{tabular}{|l|l|c|c|c|}
        \hline
        Dataset & Model & ${S}_\mathrm{reconst}$ & mIoU & ${S}_\mathrm{gen}$\\
        \hline
        \multirow{6}{*}{Crello} & AutoReg LSTM & 79.75 & 33.52 & 86.44\\
        & AutoReg Trans x1 & 85.47 & 33.86 & 87.48\\
        & AutoReg Trans x4 & 84.65 & 35.36 & 86.60\\
        \cline{2-5}
        & Oneshot LSTM & 84.95 & 40.50 & 86.85\\
        & Oneshot Trans x1 & \bf 88.67 & \bf 47.02 & 87.57\\
        & Oneshot Trans x4 & 87.75 & 45.50 & \bf 88.15\\
        \hline
        \multirow{6}{*}{RICO} & AutoReg LSTM & 87.73 & 42.51 & 93.74\\
        & AutoReg Trans x1 & \bf 94.96 & 51.13 & 94.40\\
        & AutoReg Trans x4 & 92.06 & 48.74 & 95.11\\
        \cline{2-5}
        & Oneshot LSTM & 91.01 & 51.93 & 92.05\\
        & Oneshot Trans x1 & 94.35 & \bf 60.42 & 93.90\\
        & Oneshot Trans x4 & 94.45 & 59.47 & \bf 95.14\\
        \hline
    \end{tabular}
\end{table}

For each baseline, we report the test performance of the best validation model in terms of $S_\mathrm{gen}$ that we find by a grid search over $\lambda_\mathrm{KL}$.
For other hyper-parameters, we empirically set the size of latent code $\mathbf{z}$ to 512 for Crello and 256 for RICO, and $\lambda_\mathrm{L2} = 1e-6$ in all baselines.
We train all baseline models using Adam optimizer with learning rate fixed to $1e-3$ for 500 epochs in both datasets.
For generation evaluation, we sample $\mathbf{z}$ from zero-mean unit normal distribution up to the same size to the test split.

Table \ref{tab:performance} summarizes the test evaluation metrics of the baseline models.
In Crello, oneshot Transformer x1 configuration has the best reconstruction in structure (88.67\%) and layout mIoU (47.02\%), while oneshot Transformer x4 has the best generation score (88.15\%).
In RICO, autoregressive Transformer x1 has the best structure reconstruction (94.96\%), oneshot Transformer x1 has the best mIoU (60.42\%), and oneshot Transformer x4 has the best generation score (95.14\%).

\paragraph{Performance trade-offs}
We note that the choice of $\lambda_\mathrm{KL}$ has a strong impact on the evaluation metric, and that explains the varying testing results in Table \ref{tab:performance}.
We plot in Fig~\ref{fig:performance-curve} the validation performance as we move $\lambda_\mathrm{KL}$ from $2^1$ to $2^8$ in Crello, and from $2^1$ to $2^7$ in RICO.
The plots clearly show there is a trade-off relationship between reconstruction and generation.
A smaller KL divergence indicates smoother latent space, which in turn indicates better generation quality from random sampling of $\mathbf{z}$ but hurts the reconstruction performance.
From the plots, oneshot Transformer x4 seems consistently performing well in both datasets except for $S_\mathrm{reconst}$ evaluation in RICO, where most baselines saturate the reconstruction score.
We suspect $S_\mathrm{reconst}$ saturation is due to over-fitting tendency in RICO dataset, as RICO does not contain high dimensional attributes like image feature.
Given the performance characteristics, we conjecture that oneshot Transformer performs the best.

\paragraph{Autoregressive vs. oneshot}
Table \ref{tab:performance} and Fig~\ref{fig:performance-curve} suggests oneshot models consistently perform better than the autoregressive counterparts in both datasets.
This trend makes sense, because autoregressive models cannot consider the layout placement of future elements at the current step.
In contrast, oneshot models can consider spatial relationship between elements better at inference time.

\subsection{Qualitative evaluation}

\paragraph{Reconstruction}
We compare reconstruction quality of Crello and RICO testing examples in Fig~\ref{fig:reconstruction-crello} and Fig~\ref{fig:reconstruction-rico}.
Here, we reconstruct the input deterministically by the mean latent code in Eq~\ref{eq:variational_posterior}.
Since Crello dataset has rich content attributes, we present a document visualization that fills in image and shape elements using a nearest neighbor retrieval by image features (Sec~\ref{sec:crello}), and a color map of element type bounding boxes.
For RICO, we show a color map of \emph{component} bounding boxes.

We observe that, while all baselines reasonably reconstruct documents when there are a relatively small number of elements, oneshot models tend to reconstruct better as a document becomes more complex (Second row of Fig~\ref{fig:reconstruction-crello}).

We also show how sampling $\mathbf{z}$ from $q_\phi(\mathbf{z}|X)$ results in variation in the reconstruction quality in Fig~\ref{fig:sampling-crello}.
Depending on the input, sampling sometimes leads to different layout arrangement or canvas size.

\paragraph{Interpolation}
One characteristic of VAEs is the smoothness of the latent space.
We show in Fig~\ref{fig:interpolation} an example of interpolating latent codes between two documents in Crello dataset.
The result shows a gradual transition between two documents that differ in many aspects, such as the number of elements or canvas size.
The resulting visualization is discontinuous in that categorical attributes or retrieved images change at specific point in between, but we can still observe some sense of continuity in design.

\paragraph{Generation}
Fig~\ref{fig:generation} shows randomly generated Crello design documents from oneshot Transformer x4 configuration.
Our CanvasVAE generates documents in diverse layouts and aspect ratios.
Generated documents are not realistic in that the quality is not sufficient in immediate use in real-world creative applications, but seem to already serve for inspirational purposes.
Also, we emphasize that these generated documents are easily editable thanks to the vector graphic format.

We also show in Fig~\ref{fig:generation-rico} randomly generated UIs with RICO dataset. Although sometimes generated layouts include overlapping elements, they show diverse layout arrangements with semantically meaningful structure such as toolbar or list components.

\section{Conclusion}
We present CanvasVAE, an unconditional generative model of vector graphic documents.
Our model learns an encoder and a decoder that takes vector graphic consisting of canvas and element attributes including layout geometry.
With our newly built Crello dataset and RICO dataset, we demonstrate CanvasVAE successfully learns to reconstruct and generate vector graphic documents.
Our results constitute a strong baseline for generative modeling of vector graphic documents.

In the future, we are interested in further extending CanvasVAE by generating text content and font styling, integrating pre-training of image features in an end-to-end model, and combining a learning objective that is aware of appearance, for example, by introducing differentiable rasterizer~\cite{li2020differentiable} to compute raster reconstruction loss.
We also wish to look at whether feedback-style inference~\cite{wang2018feedback} allows partial inputs such as user-specified constraints~\cite{lee2020neural}, which is commonly seen in application scenarios.

{\small
\bibliographystyle{ieee_fullname}
\bibliography{main}
}

\end{document}